\documentclass[journal]{IEEEtran}
\usepackage{amsmath,amsfonts}
\usepackage{mathrsfs}
\usepackage{algorithmic}
\usepackage{algorithm}
\usepackage{array}
\usepackage{textcomp}
\usepackage{stfloats}
\usepackage{url}
\usepackage{hyperref}
\hypersetup{
	colorlinks=true,
	linkcolor=blue,
	filecolor=blue,
	urlcolor=blue,
	citecolor=blue,
}
\usepackage{xcolor}
\usepackage{flushend}
\usepackage{verbatim}
\usepackage{graphicx}
\graphicspath{{Figures/}}
\usepackage{epstopdf}
\usepackage{cite}
\usepackage{subcaption}
\usepackage{multirow}
\usepackage{bm}
\usepackage{siunitx}
\usepackage{booktabs}
\usepackage{amssymb}
\def\BibTeX{{\rm B\kern-.05em{\sc i\kern-.025em b}\kern-.08em T\kern-.1667em\lower.7ex\hbox{E}\kern-.125emX}}

\begin{document}

\title{Semantic-Driven AI Agent Communications: Challenges and Solutions}

\author{\IEEEauthorblockN{Kaiwen Yu, Mengying Sun, Zhijin Qin, Xiaodong Xu, Ping Yang, Yue Xiao and Gang Wu}
			
\thanks{Kaiwen Yu, Ping Yang, Yue Xiao and Gang Wu are with the National Key Laboratory of Wireless Communications, University of Electronic Science and Technology of China, Chengdu 611731, China (e-mail: yukaiwen@uestc.edu.cn, yang.ping@uestc.edu.cn, xiaoyue@uestc.edu.cn, wugang99@uestc.edu.cn).}

\thanks{Mengying Sun and Xiaodong Xu are with the State Key Laboratory of Networking and Switching Technology, Beijing University of Posts and Telecommunications, Beijing, China. Xiaodong Xu is also with the Department of Broad band Communication, Peng Cheng Laboratory, Shenzhen, Guangdong, China (e-mail: smy\_bupt@bupt.edu.cn, xuxiaodong@bupt.edu.cn).}
\thanks{Zhijin Qin is with the Department of Electronic Engineering, Tsinghua University, Beijing, China. Zhijin Qin is also with the State Key Laboratory of Space Network and Communications, Beijing, China, and the Beijing National Research Center for Information Science and Technology, Beijing, China (e-mail: qinzhijin@tsinghua.edu.cn).}
}

\maketitle

\begin{abstract}
With the rapid growth of intelligent services, communication targets are shifting from humans to artificial intelligent (AI) agents, which require new paradigms to enable real-time perception, decision-making, and collaboration. Semantic communication, which conveys task-relevant meaning rather than raw data, offers a promising solution. However, its practical deployment remains constrained by dynamic environments and limited resources. To address these issues, this article proposes a semantic-driven AI agent communication framework and develops three enabling techniques. 
First, semantic adaptation transmission applies fine-tuning with real or generative samples to efficiently adapt models to varying environments. Second, semantic lightweight transmission incorporates pruning, quantization, and perception-aware sampling to reduce model complexity and alleviate computational burden on edge agents. Third, semantic self-evolution control employs distributed hierarchical decision-making to optimize multi-dimensional resources, enabling robust multi-agent collaboration in dynamic environments.
Simulation results show that the proposed solutions achieve faster convergence and stronger robustness, while the proposed distributed hierarchical optimization method significantly outperforms conventional decision-making schemes, highlighting its potential for AI agent communication networks.

\end{abstract}

\section{Introduction}
With the rapid development of intelligent technologies, communication entities are evolving from humans to artificial intelligent (AI) agents capable of environmental perception, semantic understanding, and autonomous decision-making\cite{qin2024ai}. It is expected that AI agents will be widely deployed in various fields\cite{chan2024visibility}. Typical applications include collaborative unmanned aerial vehicle (UAV) swarms, connected vehicles, and service robots. AI agent communication aims to provide global connectivity, on-demand empowerment, and controllable information exchange and task collaboration for heterogeneous agents with diverse forms, capabilities, and users\cite{du2025ai,jiang2025large}.

In contrast to traditional mobile terminals and internet of things (IoT) devices, AI agents can proactively initiate communication and collaborate to complete complex tasks assigned by humans. This places new demands on access, network connectivity, and session establishment. For example, multi-agent collaboration often requires task-driven, dynamic, and on-demand network relationships. Crucially, communications between agents is no longer constrained by human sensory or cognitive limitations. Instead of transmitting raw text, images, or audiovisual data, agents can directly exchange the semantic-level information necessary to perform tasks.

Semantic communication has emerged as a revolutionary paradigm, abandoning the traditional ``transmit first, compute later'' model in favor of a ``compute first, transmit later'' approach\cite{niu2025mathematical,getu2025semantic,sagduyu2024joint,huang2025semantic}. Its core principle is to transmit only semantic information relevant to the task, significantly reducing communication overhead and improving efficiency while ensuring task completion. Compared to traditional bit transmission, semantic communication supports cross-modal information representation and robust end-to-end transmission\cite{xia2025generative}. This technology has been recognized by 3GPP as a key enabling technology for 6G\cite{TR22870} and has proven particularly suitable for agent communication scenarios. The core of AI agents lies in ``understanding and decision-making'', which can be directly enabled through the exchange of abstract semantic features. Furthermore, semantic communication provides a unified semantic-level language structure to address the challenges of heterogeneity and cross-modal interaction between AI agents, thereby promoting efficient multi-agent collaboration and interoperability. Therefore, semantic communication could be considered a foundational technology for future AI agent communication networks.

Recent studies have demonstrated significant progress in semantic communications for text, speech, and image based on convolutional neural networks (CNNs), Transformer models, and generative models \cite{yu2025partial,chen2025communication,li2025cognitive}. However, despite the unique requirements and challenges of AI agent communication, dedicated research remains limited. Three key issues exist: (i) Agent tasks are highly dynamic and are often performed in mobile environments where wireless channels frequently fluctuate. This requires communication systems to flexibly adjust semantic goals and transmission strategies to adapt to changing task requirements and channel conditions; (ii) Edge agents face strict computational and storage constraints, making it difficult to deploy complex semantic models; and (iii) In multi-agent task collaboration, multi-dimensional resources, including communications, computation, and energy, need to self-configure and self-evolve to ensure long-term robustness.

To address these challenges, this paper investigates semantic-driven AI agent communications. We systematically analyze the fundamental issues and introduce three key enabling technologies: semantic adaptive transmission, semantic lightweight transmission, and model self-evolution. We propose a semantic-driven AI agent communication framework to capture the unique characteristics of agent interactions and address the associated challenges. Furthermore, we conduct three case studies for edge-to-edge agent communications, edge-to-base station (BS) agent communications, and multi-agent communication network scenarios. Experimental results demonstrate that the proposed strategies significantly improve task execution efficiency.

\begin{table*}[!t]
	\centering
	\renewcommand{\arraystretch}{1.5}
	\caption{Comparison between traditional UEs and AI agents in communications}
	\begin{tabular}{@{}p{3cm}p{6cm}p{6cm}@{}}
		\toprule[1.5pt]
		\textbf{Dimension} & \textbf{Traditional UEs} & \textbf{AI Agents} \\
		\midrule[0.75pt]
		Communication trigger & Passive, driven by user operations or preset rules & Proactive, autonomous access and collaboration \\
		
		Intelligence level & Limited, dependent on human sensory and control & Perception, reasoning, and decision-making \\
		
		Identity attribute & Device ID, loosely associated with user & Unique digital identity, traceable and controllable \\
		
		Capability openness & Closed, only serving as communication terminals & Open, sharing sensing, reasoning, and actuation \\
		
		Interaction content & Raw bits or multimedia data & Task-oriented semantic information and capabilities \\
		
		Network relationship & Static connections, relatively fixed topology & Dynamic networking, multi-agent collaboration \\
		\bottomrule[1.5pt]
	\end{tabular}
	\label{DifferencesFromTraditionalTerminals}
\end{table*}

\begin{figure*}[!t]	
	\centerline{\includegraphics[width=6.5in]{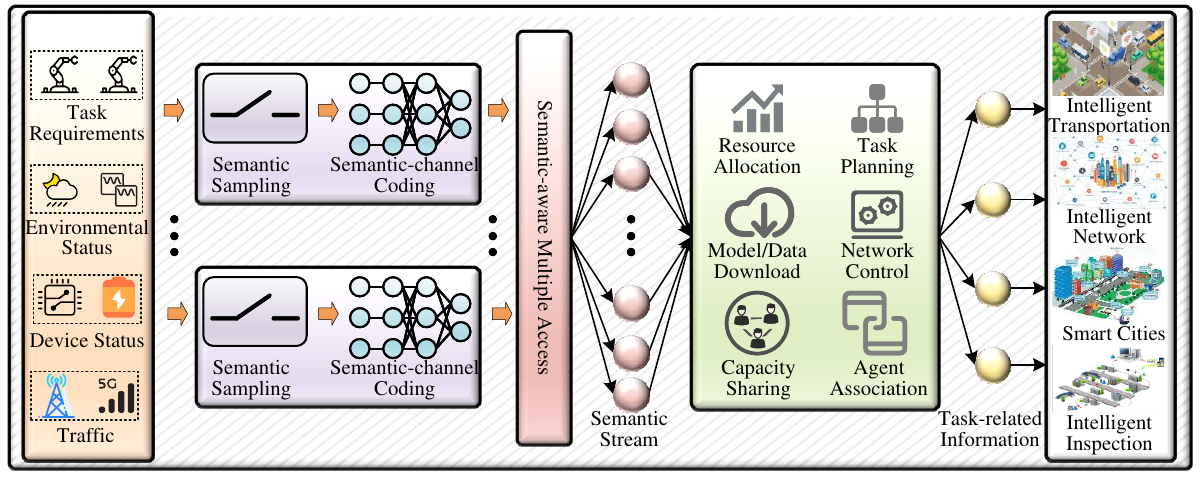}}
	\caption{Semantic-driven AI agent communication framework. The architecture illustrates three main stages: Perception-aware semantic sampling, joint semantic-channel coding, and semantic resource orchestration.}
	\label{framework}
\end{figure*}

\section{What is AI Agent Communications?}\label{s2}

\subsection{Differences From Traditional Terminals}

AI agents manifest in diverse forms, which can be categorized into three types\cite{TR22870}:

\begin{itemize}
	\item \textbf{Embodied agents}: These include humanoid robots, robotic dogs, and autonomous vehicles equipped with powerful onboard computing resources. They connect to networks via wired or wireless links and are capable of performing intelligent operations independently.
	\item \textbf{Cloud agents}: These are deployed in cloud centers, such as digital humans or AI assistants, which primarily rely on cloud-based intelligence and interact with users through communication networks.
	\item \textbf{Third-party agents}: These refer to AI-enabled devices such as smartphones, wearables, and drones that possess limited local intelligence and I/O capabilities. By leveraging collaborative integration across the device-network-cloud continuum, their capabilities can be significantly enhanced to support advanced inference and decision-making.
\end{itemize}

As shown in Table.~\ref{DifferencesFromTraditionalTerminals}, from a communication perspective, AI agents are fundamentally different from traditional entities (such as user equipment (UE)), mainly in the following aspects:

\begin{itemize}
	\item \textbf{Autonomy}: AI agents can independently perceive the environment, make decisions, and execute actions without human intervention. They are also able to proactively establish communication links with other intelligent devices.  
	\item \textbf{Digital identity}: Each AI agent is associated with a unique digital identity that is tightly linked to its owner or organization.
	\item \textbf{Capability sharing}: AI agents offer sensing, reasoning, and actuation capabilities that can be shared with others. Through capability orchestration, multiple agents can form cooperative networks to achieve collective intelligence.  
\end{itemize}

In summary, AI agent communication extends conventional human-triggered bit exchange into a new paradigm of \textit{autonomous, semantic-rich, and capability-oriented interactions among intelligent entities}, thus laying the foundation for next-generation intelligent networks.  

\begin{figure*}[!t]	
	\centerline{\includegraphics[width=7in]{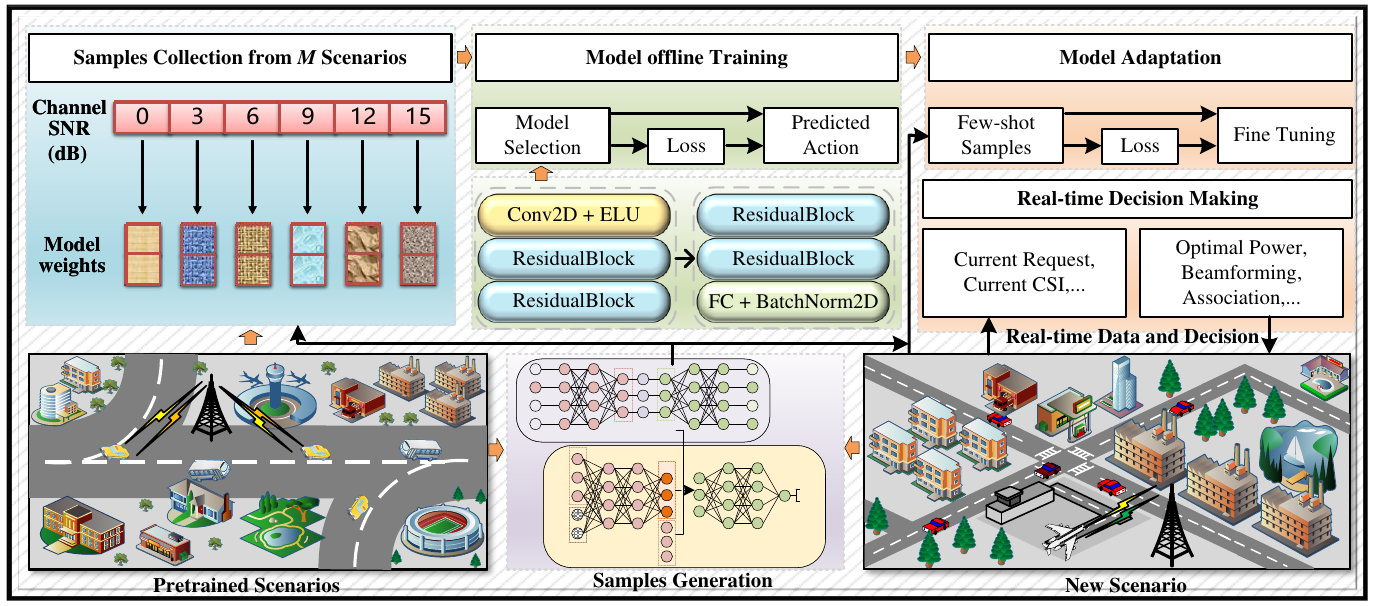}}
	\caption{Illustration of semantic adaptation transmission for edge-to-edge agent communications. The semantic transmission model is pre-trained with multiple scenarios to learn the policies. In real-time deployment, the model quickly adapts to the new scenario by fine-tuning the pre-trained model.}
	\label{ModelAdaptive}
\end{figure*}

\subsection{Semantic-driven Agent Communication Frameworks}

As illustrated in Fig.~\ref{framework}, in semantic-driven AI agent communications, the end-to-end process involves three key stages: semantic sampling, joint semantic-channel coding, and semantic resource orchestration.

\begin{itemize}
	\item \textbf{Perception-aware semantic sampling}: As the first step in the perception-cognition-decision-learning-action cycle, semantic sampling defines the quality of information that supports downstream intelligence. Agents acquire multimodal data from their surroundings through cameras, microphones, and industrial sensors. Since raw data is typically redundant and unstructured, agents must extract task-relevant information while filtering out noise and irrelevant values, thereby improving communication efficiency and ensuring meaningful inputs for cognition and decision-making. Moreover, by integrating performance monitoring and feedback mechanisms, agents can evaluate the effectiveness of their sampling strategies and adapt them through self-learning\cite{yu2025partial}, gradually evolving toward more autonomous and efficient communication behaviors.

	\item \textbf{Joint semantic–channel coding}: Semantic compression combined with channel optimization ensures efficient and robust transmission. Different from the traditional design of separate source and channel coding, joint semantic-channel coding utilizes neural networks (NN) to jointly learn the semantic features of the source and the statistical properties of the channel. By dynamically allocating codewords based on channel conditions (e.g., high vs. low signal-to-noise ratio (SNR)), the system balances transmission efficiency with robustness against interference.
	
	\item \textbf{Semantic resource orchestration}: For collaborative tasks, agents form temporary sub-networks that are dynamically created based on task demands and disbanded after the task is completed. Within these subnetworks, agents semantically describe and share their capabilities, interpret user intent, and coordinate planning, offloading, scheduling, and execution. Semantic-level orchestration goes beyond data exchange to enable functional complementarity between agents, fostering collaborative intelligence and improving overall task efficiency.
\end{itemize}

Overall, this three-stage process enables AI agent communication to evolve from primitive data exchange to meaning-centered, task-oriented, and resource-aware interactions, laying the foundation for the next generation of intelligent networks.

\subsection{Challenges in Semantic-Driven Agent Communications}

\textbf{Challenges in semantic dynamic variation:} The highly dynamic nature of agent tasks requires communication systems to adapt semantic goals based on changing task and environmental conditions. In rapidly changing wireless channels, performance often degrades due to the mismatch between training and deployment environments, driven by mobility, multipath fading, and unknown interference. Developing end-to-end semantic transmission schemes that can dynamically adapt to different tasks and channels, while establishing theoretical models and performance bounds, remains a key research challenge.

\textbf{Challenges in semantic model complexity:} Agents often operate under strict computational and energy constraints. This is especially true for physical agents such as drones, robots, and extended reality (XR) devices, which cannot support large-scale models containing billions of parameters. Relying solely on cloud resources introduces latency and congestion in multi-user scenarios, hindering real-time responsiveness. Furthermore, agent sensing is often limited to a partial view of the environment. Given that semantic communication trades computational complexity for transmission efficiency, designing lightweight, edge-oriented semantic communication architectures is crucial for balancing computational overhead and communication efficiency.

\textbf{Challenges in multi-agent collaboration:} In contrast to traditional devices that passively follow user commands, agents proactively initiate communication and collaboration to complete human-assigned tasks. These interactions are frequent, complex, and highly dynamic. As user and task demands continually change, resource allocation strategies must be adjusted in real time, posing significant challenges to the scalability and long-term evolution of semantic-driven agent communication networks.

\begin{figure*}[!t]	
	\centerline{\includegraphics[width=6in]{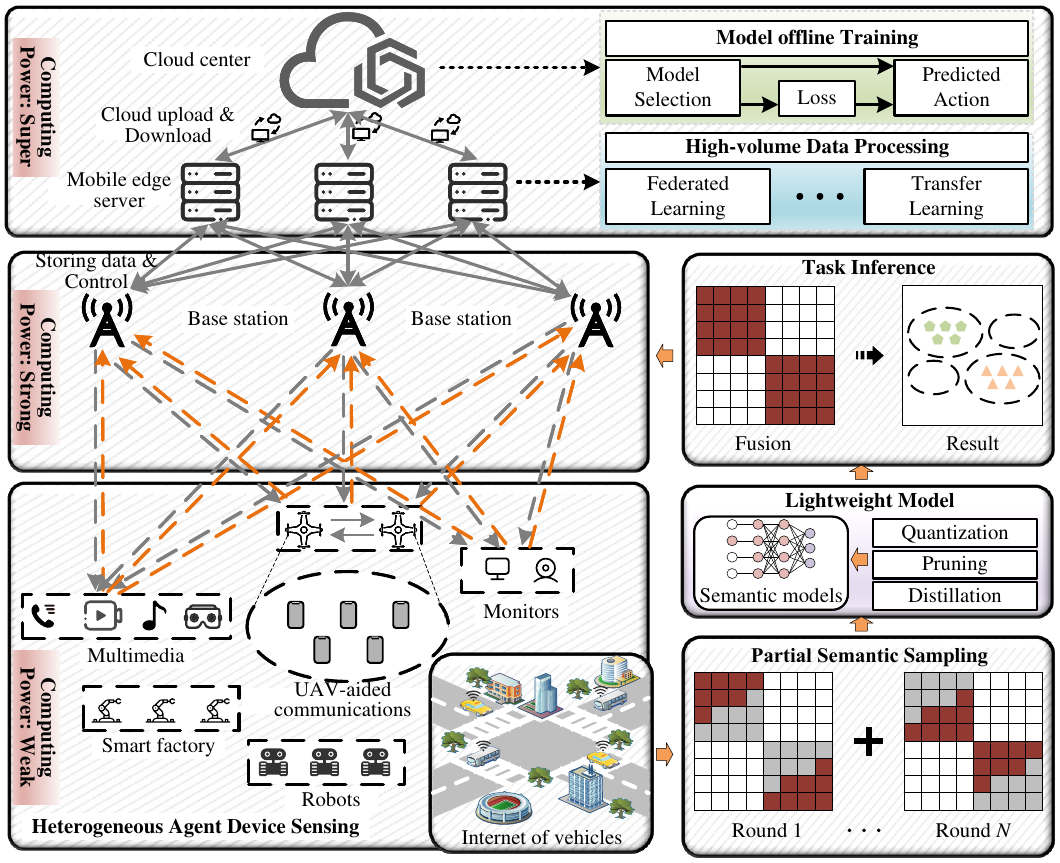}}
	\caption{Illustration of lightweight semantic transmission for edge-to-BS agent communications. Edge agents are constrained by limited computing power and bandwidth, while BS agents generally have greater computational resources, with cloud agents providing the super capacity. The architecture integrates parameter quantization, pruning, and distillation to reduce the complexity and computational demands of semantic models. In addition, partial sample processing is employed to further relieve the burden on edge agents.}
	\label{ModelLightweight}
\end{figure*}

\section{Key Technologies for Semantic-Driven AI Agent Communications}
To address aforementioned challenges, we propose the following three key technologies.
\subsection{Semantic Adaptive Transmission via Fine-Tuning}

In edge-to-edge agent communications, the environments on both sides are dynamically changing. Online fine-tuning of semantic model involves continuously updating the parameters of the pre-trained semantic codec during the operation process based on changing channel conditions and task requirements. Compared to traditional offline training, online fine-tuning avoids retraining large models from scratch, enabling rapid adaptation with reduced computational and communication overhead.

As shown in Fig.~\ref{ModelAdaptive}, in new scenarios, the channel states and communication tasks may differ from those in the pre-trained environment. Retraining the model from scratch is costly. To address this issue, a fine-tuning strategy can be applied to update the model using only a small number of new samples or data synthesized by the generative models\cite{yu2024two}. Generative adversarial networks (GAN), in particular, have been widely validated for their ability to capture complex distributions and generate diverse data while preserving the statistical properties of the original dataset\cite{yu2024two}. By fine-tuning the semantic codec with GAN-generated samples, the system achieves more efficient weight adaptation, thereby improving robustness and adaptability under dynamic environments.

By combining offline training with online fine-tuning, the semantic-driven AI agent communication framework can continuously refine its model, achieving rapid policy adjustment and enhanced task adaptability in complex and dynamic environments.

\begin{figure*}[!t]	
	\centerline{\includegraphics[width=7in]{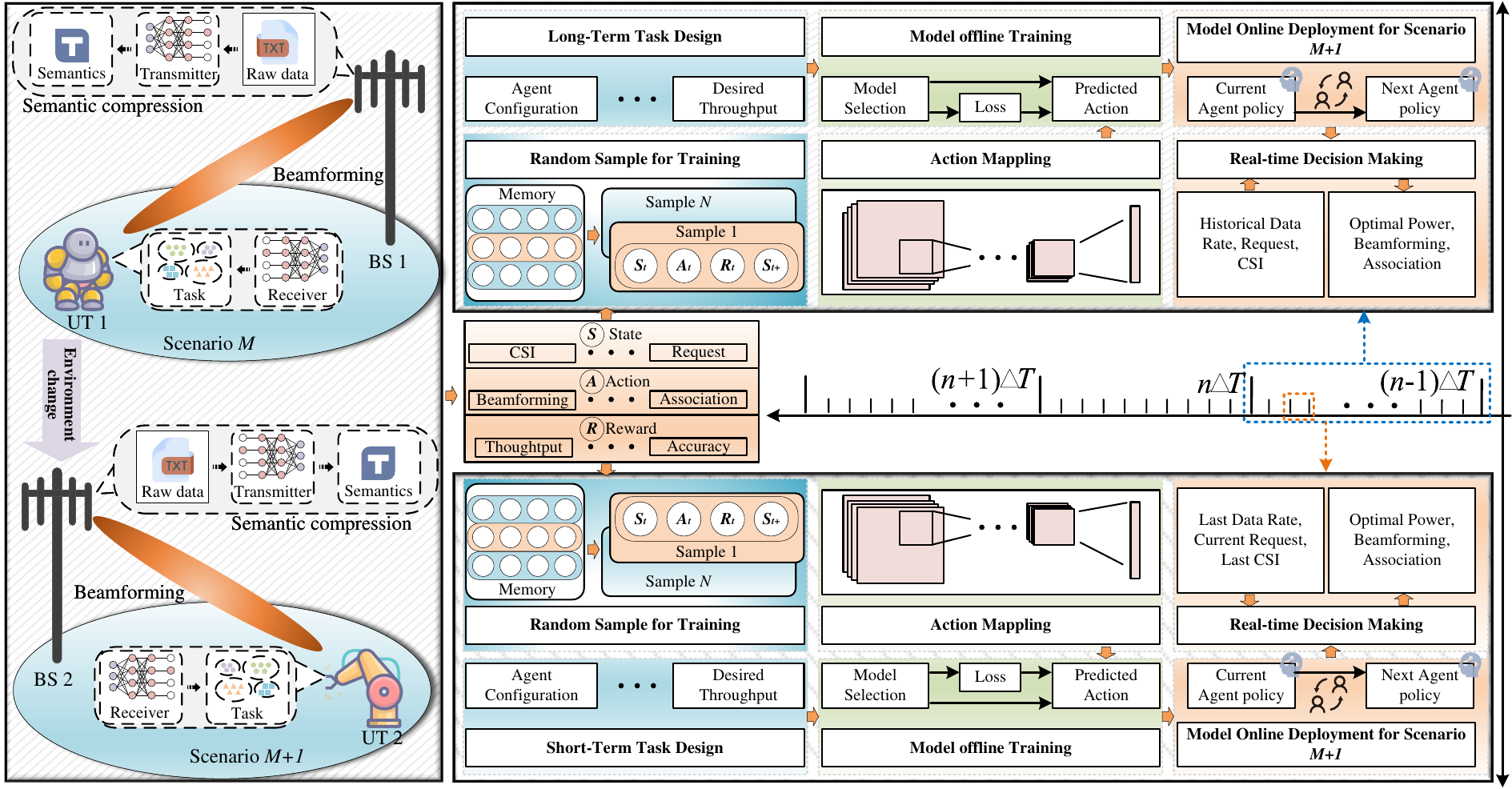}}
	\caption{Illustration of distributed hierarchical multi-agent reinforcement learning for multi-agent communication networks. This framework can be defined based on different communication tasks. Agents gradually refine their action configurations based on high-level and low-level strategies to improve the overall network's transmission performance. By managing and coordinating optimization objectives, the network is able to self-configure, self-optimize, and adapt to dynamic environments and semantic requirements.}
	\label{DHC}
\end{figure*}

\subsection{Lightweight Semantic Transmission}

In edge-to-BS agent communications, semantic encoders are typically deployed on mobile agents to extract task-relevant information for transmission. However, the limited memory, computing resources, and energy of edge devices pose significant challenges for deploying advanced semantic models. Consequently, lightweight design strategies are essential to enable efficient semantic transmission under such resource constraints.

The computational complexity of the semantic codec depends on the modality of the data to be transmitted. For simple bit-level data, a few fully connected layers are sufficient. By contrast, more complex data types, such as text or images, require advanced NNs. While complex NNs yield better performance, their deployment in edge agents is constrained by computation and memory costs. As illustrated in Fig.~\ref{ModelLightweight}, lightweight techniques such as parameter pruning and quantization can be applied to semantic codecs. Pruning reduces network size by eliminating redundant parameters, where insensitive neurons or layers are discarded with minimal impact on performance. Quantization further reduces complexity by converting weights and activations from floating-point numbers to low-precision integers, thereby lowering memory usage and simplifying computation.

In addition to model compression, redesigning the communication framework can also reduce the burden on individual edge agents. As shown in Fig.~\ref{ModelLightweight}, the partial sampling enables edge devices to extract and transmit only task-related portions of semantic information rather than processing the complete raw data\cite{yu2025partial}. In this paradigm, multiple lightweight agents capture different aspects of the target environment and collect partial data for encoding and transmission to the BS for fusion and inference. For example, multiple agents equipped with low-resolution cameras can each sample fragments of the scene and forward them to a receiver, which reconstructs the information required to complete the task through semantic fusion. The partial sampling not only reduces the computational requirements of individual agents but also allows the system to maximize task relevance by aggregating different semantic contributions.

\subsection{Semantic Self-Evolution}

In multi-agent communication networks, information transmission is highly complex, resource scheduling involves multiple dimensions, and both the communication environment and task requirements may change dynamically. To address these challenges, we propose a distributed multi-timescale hierarchical deep reinforcement learning framework, as shown in Fig.~\ref{DHC}. Within this framework, each agent device is equipped with autonomous resource decision-making capabilities. On large timescales, it dynamically adjusts physical layer parameters such as power allocation and beamforming to ensure stable transmission under varying channel conditions. On small timescales, it adaptively manages semantic extraction and compression strategies to maximize task relevance and efficiency. Through distributed hierarchical control, agents collaborate to achieve resource self-configuration and enable continuous self-evolution of the system.

During training, the hierarchical framework supports continuous optimization and coupled updates of the two-layer decision-making strategies. Specifically, the large-timescale strategy is first updated to suppress inter-channel interference and enhance link robustness. Building on this stable wireless environment, the small-timescale strategy is then optimized to improve semantic compression efficiency and task execution performance. By jointly updating the two NNs, the framework achieves adaptive multi-dimensional resource optimization, ensuring that the system continuously evolves towards optimal performance under dynamic network conditions.

\section{Case Study and Simulation Evaluation}

In this section, we conduct detailed simulations on three case studies to evaluate the performance of our proposed solutions.

\subsection{Case Study for Edge-to-Edge Agent Communications}
We first consider an edge-to-edge agent communication scenario, where each agent is equipped with a semantic codec and a conditional GAN-based channel simulator. Performance is measured in terms of peak signal-to-noise ratio (PSNR) and mean squared error (MSE) over an AWGN channel. To evaluate the effectiveness of real-time fine-tuning, the channel SNR between agents is gradually reduced in 3 dB increments, simulating worsening channel conditions as the inter-agent distance increases-from 21 dB to 9 dB in four steps.
\begin{figure}[!t]
	\centerline{\includegraphics[width=2.8in]{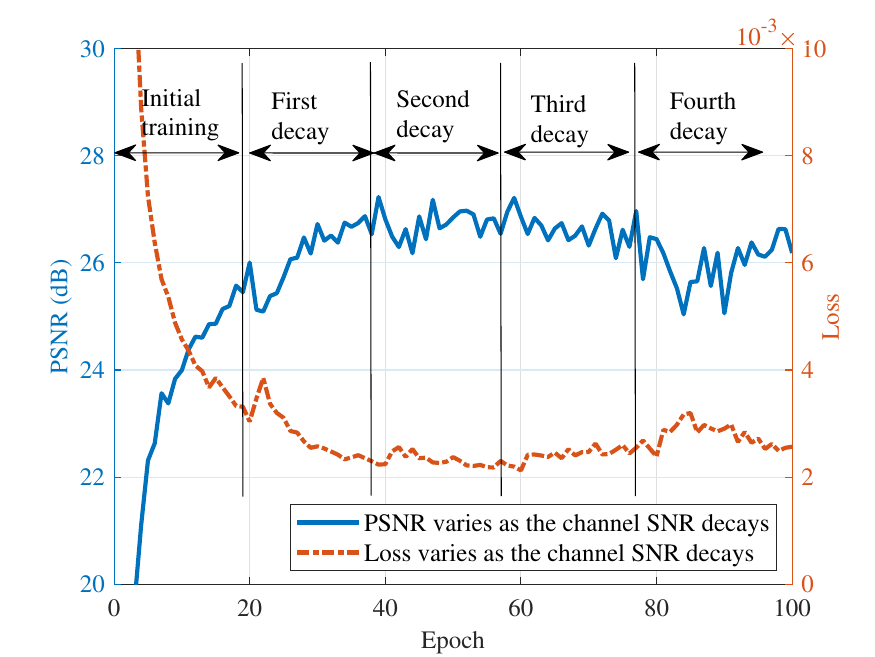}}
	\caption{Performance and convergence speed of proposed semantic real-time fine-tuning in edge-to-edge agent communications.}
	\label{PSNR_awgn}
\end{figure}
As shown in Fig.~\ref{PSNR_awgn}, the framework rapidly adapts to dynamic channel conditions. During the initial training phase (epochs 0–20), the MSE loss decreases while PSNR increases, demonstrating rapid adaptation. Each decrease in the SNR temporarily impacts performance; however, the system stabilizes within 2-3 epochs. Over time, the impact of SNR fluctuations decreases, demonstrating the robustness and reliability of the framework under varying channel conditions.

\subsection{Case Study for Edge-to-BS Agent Communications}
We further consider the edge-to–BS agent communication scenario, where edge agents are constrained by limited computing power and bandwidth and can only transmit partial content, while BS agents typically possess more powerful computing resources. Under the guidance of the receiver, the edge agent selectively sample key semantic content relevant to the communication task and gradually transmit it. Taking an image transmission task as an example, the edge agent selectively samples small-size semantic features from a large source image and transmits them to the BS agent. Through multiple sampling and transmission cycles, the receiver successfully completes the communication task. Classification accuracy is used as the evaluation metric.
\begin{figure}[!t]
	\centerline{\includegraphics[width=2.8in]{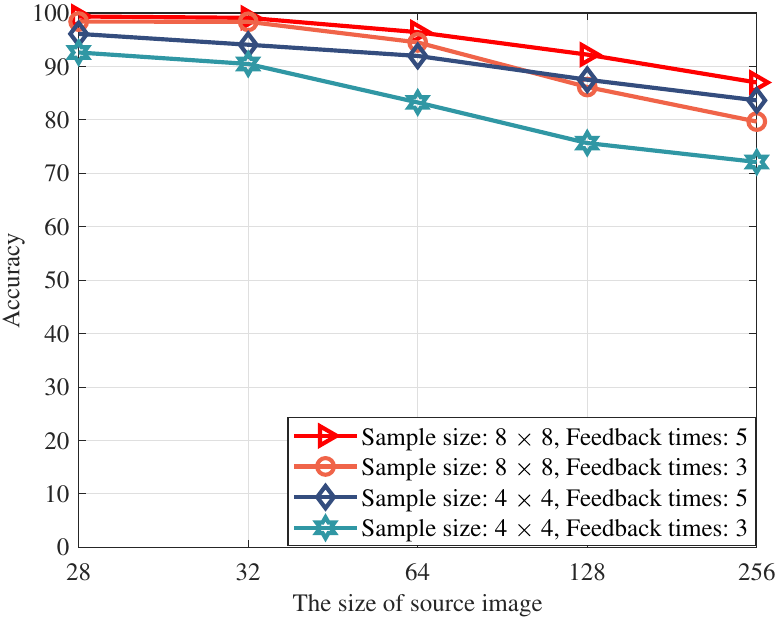}}
	\caption{Accuracy of proposed partial sampling scheme in edge-to-BS agent communications.}
	\label{Accuracy_awgn}
\end{figure}
Fig.~\ref{Accuracy_awgn} shows the task performance of the proposed partial sampling scheme under AWGN channel for different source sizes, sampling sizes and feedback round numbers. With fixed sampling and feedback, accuracy decreases with increasing source size because capturing key semantics becomes more challenging. Conversely, increasing the sampling size allows the edge agent to capture more semantic information, thereby improving accuracy. The impact of the number of feedback rounds is most significant when the sampling size is $4 \times 4$ and weakens when the sampling size is $8 \times 8$, especially for larger sources. Overall, larger sampling sizes and more feedback rounds provide the receiver with richer information, thereby improving task performance.

\subsection{Case Study for Multi-Agent Communication Networks}
At last, we consider a multi-agent communication networks where inter-agent interference plays a critical role. The network operates over consecutive time slots, with the large timescale spanning multiple time slots and the small timescale corresponding to a single time slot. On the large timescale, the BS agents collaborate to optimize beamforming and power allocation to mitigate inter-agent interference. On the small timescale, the BS agents dynamically adjust the semantic compression configuration to optimize link-level semantic transmission. The large timescale state includes the current equivalent channel gain, total interference noise power, and previous power and beam operations. The small timescale state includes the previous semantic compression operation, the current channel gain, and interference noise power. Accordingly, the large timescale operation consists of the beam codebook and power codebook index, while the small timescale operation consists of the semantic compression codebook index. The reward is defined as the quality of experience (QoE) gain. For specific parameter settings, please refer to \cite{yu2025distributed}.
\begin{figure}[!t]
	\centerline{\includegraphics[width=2.8in]{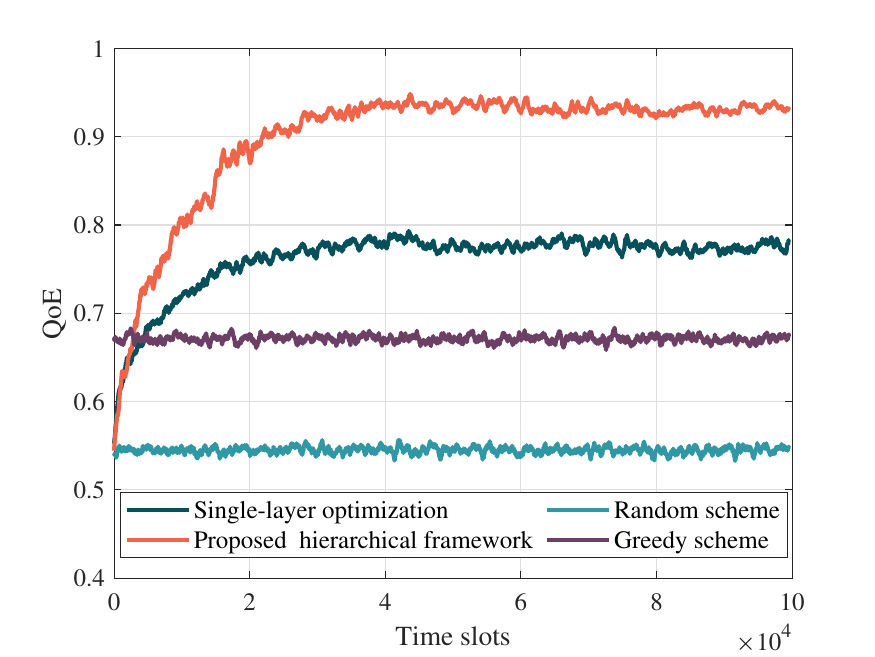}}
	\caption{QoE score vs. Time slots under different schemes in multi-agent communication networks.}
	\label{QoEvsTimeslots}
\end{figure}
As shown in Fig.~\ref{QoEvsTimeslots}, the proposed framework initially exhibits a low QoE score but gradually achieves stable performance by learning the wireless environment and communication task. Furthermore, the proposed approach significantly outperforms other traditional single-layer methods, demonstrating its ability to adapt to dynamic environments and adjust strategies in real time to effectively meet task requirements.

\section{Conclusion}

In this paper, we have explored the challenges faced by semantic-driven AI agent communications and proposed innovative frameworks to address them. This framework integrates core technologies such as semantic adaptive transmission, semantic lightweight transmission, and semantic self-evolution, laying a practical foundation for realizing semantic communication in AI agent networks. Simulation results demonstrate that the frameworks achieve superior convergence speed and overall performance compared with traditional approaches. Our approach is not limited to the discussed of AI agent communications but also offers a promising approach for developing scalable solutions for other scenarios, such as edge intelligence and smart internet of vehicles.

\bibliography{Refer_Mag_SCAgent_SC}
\bibliographystyle{IEEEtran}

\end{document}